\documentclass[10pt,twocolumn,letterpaper]{article}

\usepackage{cvpr}
\usepackage{times}
\usepackage{epsfig}
\usepackage{graphicx}
\usepackage[fleqn]{amsmath}
\usepackage{mathtools}
\usepackage{amssymb}
\usepackage{booktabs}
\usepackage{array}
\usepackage{siunitx}
\usepackage[inline]{enumitem}
\usepackage{colortbl}
\usepackage[export]{adjustbox}
\usepackage{xcolor}
\usepackage{microtype}
\usepackage{flushend}
\usepackage[font=footnotesize,skip=0pt]{caption}

\newcolumntype{P}[1]{>{\centering\arraybackslash}p{#1}}

\DeclareSIUnit{\million}{M}
\DeclareSIUnit{\billion}{B}

\renewcommand{\eqref}[1]{Eq.~(\ref{#1})}
\newcommand{\figref}[1]{Fig.~\ref{#1}}
\newcommand{\tabref}[1]{Tab.~\ref{#1}}

\newcommand{\nset}[1]{\mathbb{#1}}
\newcommand{\myforall}{\kern.05em\ensuremath\forall\kern-.63em\rotatebox{110}{\rule{.73em}{.4pt}}}

\usepackage[pagebackref=true,breaklinks=true,letterpaper=true,colorlinks,bookmarks=false]{hyperref}

\cvprfinalcopy % *** Uncomment this line for the final submission

\def\cvprPaperID{29} % *** Enter the CVPR Paper ID here

\ifcvprfinal\fi
\begin{document}

%%%%%%%%% TITLE
\title{MOPT: Multi-Object Panoptic Tracking}

\author{Juana Valeria Hurtado \qquad Rohit Mohan \qquad Wolfram Burgard \qquad Abhinav Valada\\
University of Freiburg}

\twocolumn[{%
\renewcommand\twocolumn[1][]{#1}%
\maketitle
\begin{center}
\centering
\setlength{\tabcolsep}{0.1cm}
\setlength{\fboxsep}{0pt}
{\renewcommand{\arraystretch}{1.5}% for the vertical padding
\vspace{-0.7cm}
\begin{tabular}{p{5.65cm} p{5.65cm} p{5.65cm}}
\includegraphics[width=\linewidth]{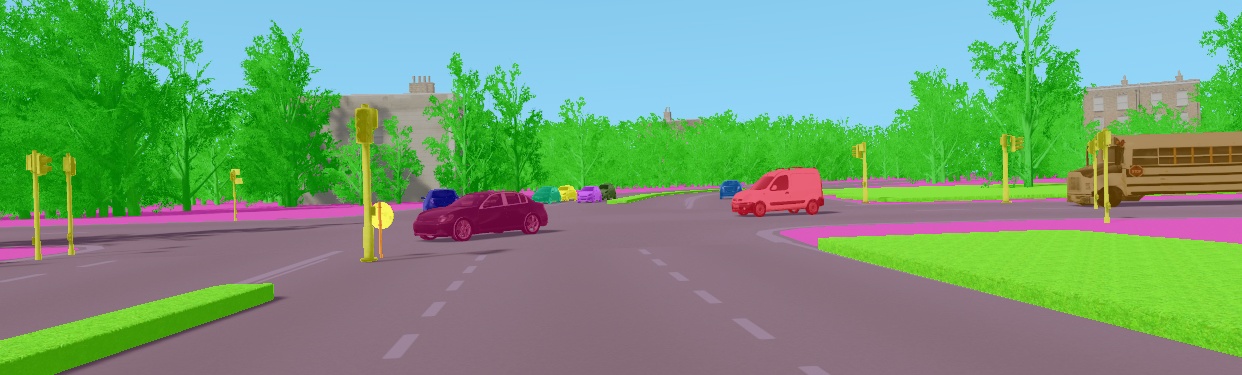} &
\includegraphics[width=\linewidth]{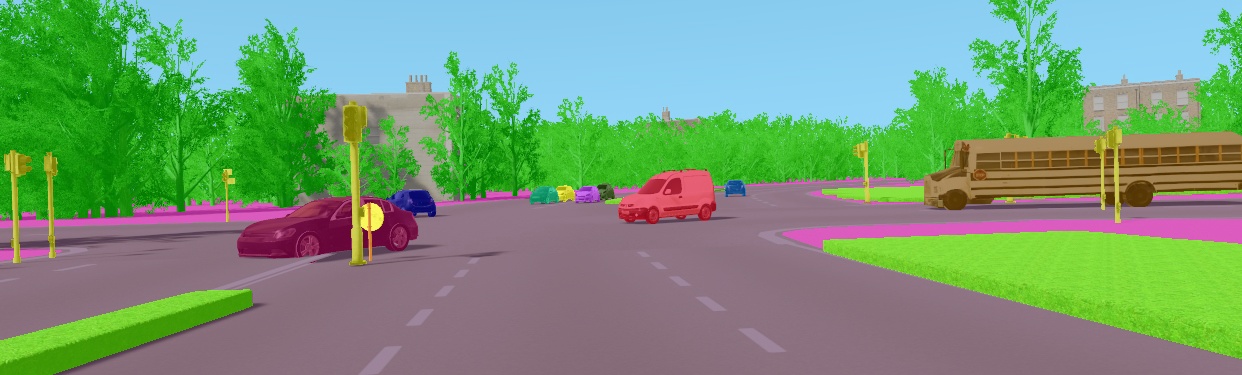} & 
\includegraphics[width=\linewidth]{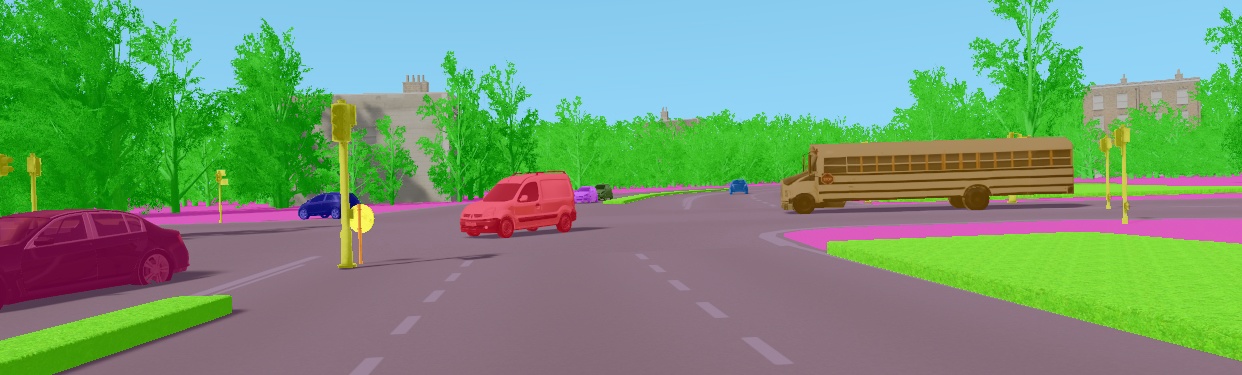} \\
\includegraphics[width=\linewidth, cfbox=lightgray 0.4pt 0pt 0pt]{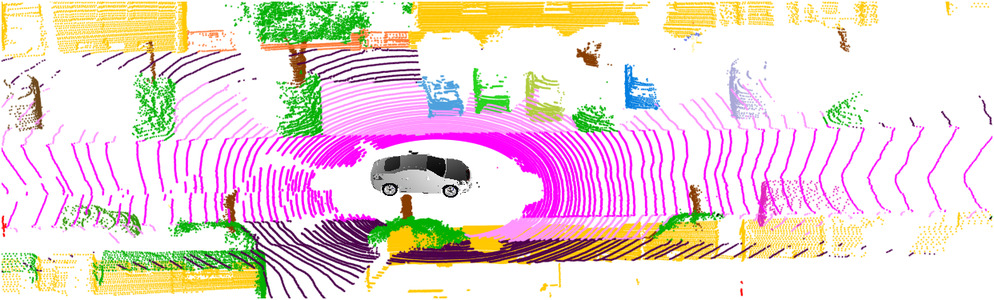} &
\includegraphics[width=\linewidth, cfbox=lightgray 0.4pt 0pt 0pt]{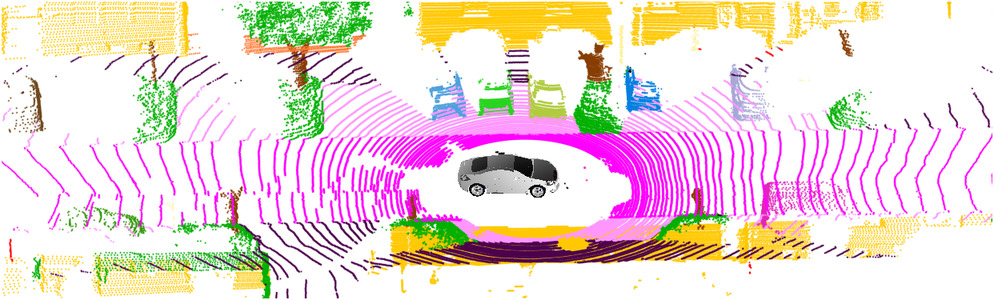} &
\includegraphics[width=\linewidth, cfbox=lightgray 0.4pt 0pt 0pt]{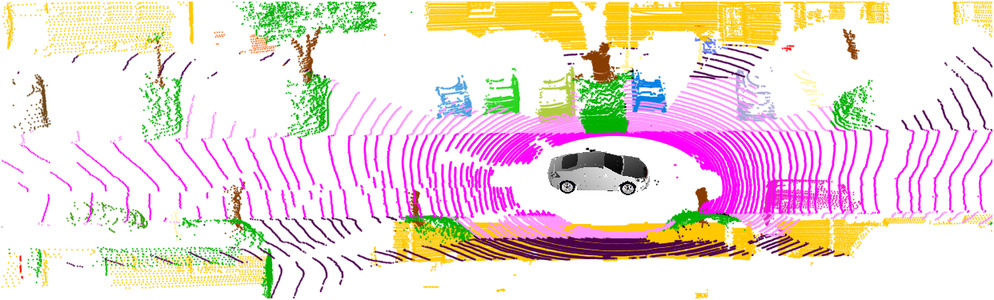} \\
\end{tabular}}
\footnotesize
\captionof{figure}{MOPT output overlaid on the image\textbackslash{LiDAR} input from the Virtual KITTI 2 (top row) and SemanticKITTI (bottom row) datasets. MOPTS unifies semantic segmentation, instance segmentation and multi-object tracking to yield segmentation of ‘stuff’ classes and segmentation of ‘thing’ classes with temporally consistent instance IDs. Observe that the tracked ‘thing’ instances retain their color-code temporally in subsequent timesteps.}\label{fig:datasetEx}
\end{center}%
}]
%\thispagestyle{empty}

%%%%%%%%% ABSTRACT
\begin{abstract}
\vspace{-0.1cm}
Comprehensive understanding of dynamic scenes is a critical prerequisite for intelligent robots to autonomously operate in their environment. Research in this domain, which encompasses diverse perception problems, has primarily been focused on addressing specific tasks individually rather than modeling the ability to understand dynamic scenes holistically. In this paper, we introduce a novel perception task denoted as multi-object panoptic tracking (MOPT), which unifies the conventionally disjoint tasks of semantic segmentation, instance segmentation, and multi-object tracking. MOPT allows for exploiting pixel-level semantic information of ‘thing’ and ‘stuff’ classes, temporal coherence, and pixel-level associations over time, for the mutual benefit of each of the individual sub-problems. To facilitate quantitative evaluations of MOPT in a unified manner, we propose the soft panoptic tracking quality (sPTQ) metric. As a first step towards addressing this task, we propose the novel PanopticTrackNet architecture that builds upon the state-of-the-art top-down panoptic segmentation network EfficientPS by adding a new tracking head to simultaneously learn all sub-tasks in an end-to-end manner. Additionally, we present several strong baselines that combine predictions from state-of-the-art panoptic segmentation and multi-object tracking models for comparison. We present extensive quantitative and qualitative evaluations of both vision-based and LiDAR-based MOPT that demonstrate encouraging results.
\vspace{-0.3cm}
\end{abstract}

%%%%%%%%% BODY TEXT
\section{Introduction}

Comprehensive scene understanding is a critical challenge that requires tackling multiple tasks simultaneously to detect, localize, and identify the scene elements as well as to understand the occurring context, dynamics, and relationships. These fundamental scene comprehension tasks are crucial enablers of several diverse applications~\cite{sa2018weedmap, valada2016convoluted, schutt2019semantic, valada2017deep, wojek2010monocular} including autonomous driving, robot navigation, augmented reality and remote sensing. Typically, these problems have been addressed by solving distinct perception tasks individually, i.e., image\textbackslash{pointcloud} recognition, object detection and classification, semantic segmentation, instance segmentation, and tracking. The state of the art in these tasks have been significantly advanced since the advent of deep learning approaches, however their performance is no longer increasing at the same groundbreaking pace~\cite{aiindex2019}. Moreover, as most of these tasks are required to be performed simultaneously in real-world applications, the scalability of employing several individual models is becoming a limiting factor. In order to mitigate this emerging problem, recent works~\cite{kirillov2018panoptic,valada2018incorporating,voigtlaender2019mots,radwan2018multimodal} have made efforts to exploit common characteristics of some of these tasks by jointly modeling them in a coherent manner.

Two such complementary tasks have gained a substantial amount of interest in the last few years due to the availability of public datasets~\cite{behley2020benchmark,cabon2020virtual} and widely adopted benchmarks~\cite{voigtlaender2019mots,behley2020benchmark}. The first of which is panoptic segmentation~\cite{kirillov2018panoptic} that unifies semantic segmentation of ‘stuff’ classes which consist of amorphous regions and instance segmentation of ‘thing’ classes which consist of countable objects. While the second task is Multi-Object Tracking and Segmentation (MOTS)~\cite{voigtlaender2019mots} which extends multi-object tracking to the pixel level by unifying with instance segmentation of ‘thing’ classes. Since the introduction of these tasks, considerable advances have been made in both panoptic segmentation~\cite{kirillov2019panoptic,xiong2019upsnet, porzi2019seamless,mohan20epsn} and MOTS~\cite{porzi2019learning, luiten2020unovost, luiten2019video, wang2019fast} which has significantly improved the performances of the previously saturating sub-tasks. Motivated by this observation, we aim to further push the boundaries by unifying panoptic segmentation and MOTS, i.e., interconnecting semantic segmentation, instance segmentation, and multi-object tracking into a holistic scene understanding problem.

A straightforward approach to tackle this unified task would be to combine the predictions of task-specific networks in a post-processing step. However, this introduces additional complexities since the overall performance largely depends on the capabilities of the individual networks that have no way of influencing the performance of their complementary task counterpart. This implies that in our case, the tracking inference directly relies on the instance segmentation performance, which again relies on the pixel-level fused ‘thing’ predictions from panoptic segmentation. Moreover, such disjoint networks also ignore supplementary cues for object recognition and segmentation provided by the temporal consistency of object identities across frames. More importantly, it entails running more number of networks in parallel which increases the overall computational complexity, thereby limiting their adoption for real-world applications due to the lack of scalability.

In this paper, we introduce a new perception task that we name Multi-Object Panoptic Tracking (MOPT). MOPT unifies the distinct tasks of semantic segmentation (pixel-wise classification of ‘stuff’ and ‘thing’ classes), instance segmentation (detection and segmentation of instance-specific ‘thing’ classes) and multi-object tracking (detection and association of ‘thing’ classes over time) as demonstrated in \figref{fig:datasetEx}. The goal of this task is to encourage holistic modeling of dynamic scenes by tackling problems that are typically addressed disjointly in a coherent manner. Additionally, we present the PanopticTrackNet architecture, a single end-to-end learning model that addresses the proposed MOPT task. The proposed architecture consists of a shared backbone with the 2-way Feature Pyramid Network (FPN)~\cite{mohan20epsn}, three task-specific heads, and a fusion module that adaptively computes the multi-object panoptic tracking output in which the number of tracked ‘thing’ classes per image could vary. Furthermore, we present several simple baselines for the MOPT task by combining predictions from disjoint state-of-the-art panoptic segmentation networks with multi-object tracking methods. To facilitate quantitative performance evaluations, we propose the soft Panoptic Tracking Quality (sPTQ) metric that extends the standard Panoptic Quality (PQ) metric to account for ‘thing’ masks that were incorrectly tracked. We present extensive experimental results using two different modalities, vision-based MOPT and LiDAR-based MOPT on the challenging Virtual~KITTI~2~\cite{cabon2020virtual} and SemanticKITTI~\cite{behley2020benchmark} datasets respectively. With our findings, we demonstrate the feasibility of training MOPT models without restricting or ignoring the input dynamics and providing useful instance identification and semantic segmentation that are also coherent in time.

In summary, the primary contributions of this paper are:
\begin{itemize}[noitemsep]
  \item We introduce the new task of Multi-Object Panoptic Tracking (MOPT) that unifies semantic segmentation, instance segmentation, and multi-object tracking into a single coherent dynamic scene understanding task.
  \item We propose the novel PanopticTrackNet architecture that consists of a shared backbone with task-specific instance, semantic, and tracking heads, followed by a fusion module that yields the MOPT output.
  \item We present several simple yet effective baselines for the MOPT task.
  \item We propose the soft Panoptic Tracking Quality (sPTQ) metric that jointly measures the performance of ‘stuff’ segmentation, ‘thing’ detection and segmentation, and ‘thing’ tracking.
  \item We present quantitative and qualitative results of both vision-based and LiDAR-based MOPT using our proposed PanopticTrackNet.
  \item We make our code and models publicly available at \url{http://rl.uni-freiburg.de/research/panoptictracking}.
\end{itemize}

%%%%%%%%%%%%%%%%%%%%%%%%%%%%%%%%%%%%%%%%%%%%%%%%%%%%%%%%%%%%%%%%%%%%%%%%%%%%%%%%
\section{PanopticTrackNet Architecture}

\begin{figure*}
  \centering
  \includegraphics[width=0.8\textwidth]{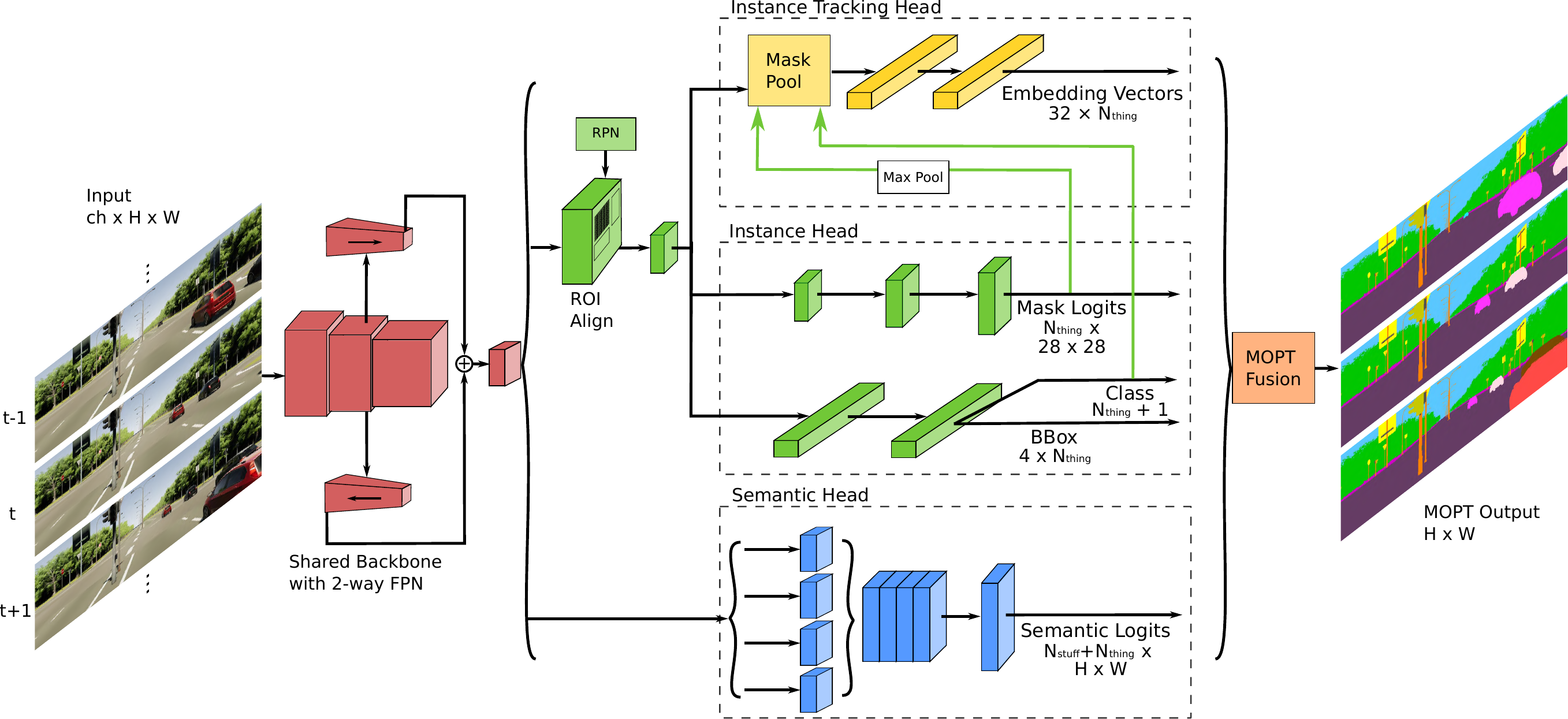}
  \vspace{5pt}
  \caption{Overview of our proposed PanopticTrackNet architecture. Our network consists of a shared backbone with the 2-way FPN (red), semantic segmentation head (blue), instance segmentation head (green), instance tracking head (yellow), and the MOPT fusion module. The fusion module adaptively combines the predictions from each of the aforementioned heads to simultaneously yield pixel-level predictions of ‘stuff’ classes and instance-specific ‘thing’ classes with temporally tracked instance IDs. Our entire network is trained in an end-to-end manner to learn these three tasks in a coherent manner.}
  \label{fig:arch}
  \vspace{-0.2cm}
\end{figure*}

The goal of our proposed architecture illustrated in \figref{fig:arch} is to assign a semantic label to each pixel in an image, an instance ID to ‘thing’ classes, and a tracking ID to each object instance thereby incorporating temporal tracking of object instances into the panoptic segmentation task. We build upon the recently introduced state-of-the-art EfficientPS~\cite{mohan20epsn} architecture for panoptic segmentation. To this end, we employ a novel shared backbone with the 2-way FPN to extract multi-scale features that are subsequently fed into three task-specific heads that simultaneously perform semantic segmentation, instance segmentation, and multi-object tracking. Finally, we adaptively fuse the task-specific outputs from each of the heads in our fusion module to yield the panoptic segmentation output with temporally tracked instances. 

\subsection{Shared Backbone}

The shared backbone that we employ is based on the EfficientNet-B5~\cite{tan2019efficientnet} topology with the 2-way Feature Pyramid Network (FPN)~\cite{mohan20epsn}. This combination enables bidirectional flow of information during multi-scale feature aggregation and yields features at four different resolutions with 256 filters each, namely downsampled by $\times4$, $\times8$, $\times16$, and $\times32$ with respect to the input. We make two main changes to the standard EfficientNet architecture. First, we remove the classification head as well as the Squeeze-and-Excitation connections as they were found to suppress the localization ability. Second, we replace the batch normalization layers with synchronized Inplace Activated Batch Normalization (iABN sync)~\cite{rota2018place} followed by a LeakyReLU activation function. This allows synchronization across different GPUs during multi-GPU training and leads to positive effects on model convergence in addition to conserving GPU memory while computing in-place operations. This backbone has been demonstrated to achieve a good trade-off between performance and computational complexity compared to other widely used backbone networks~\cite{mohan20epsn}.

\subsection{Semantic Segmentation Head}

We employ a three-module semantic segmentation head~\cite{mohan20epsn} that captures fine features and long-range context while mitigating their mismatch in an effective manner. To this end, first, the outputs of the shared backbone are separated into large-scale (downsampled by $\times4$ and $\times8$) and small-scale features (downsampled by $\times16$ and $\times32$). The large-scale features are taken as the input to the Large Scale Feature Extractor (LSFE) module that consists of two cascaded $3 \times 3$ separable convolutions with 128 output filters. Simultaneously, the small-scale features are each fed into two parallel Dense Prediction Cells (DPC)~\cite{chen2018searching}. The last module in the semantic head performs mismatch correction between the large-scale and small-scale features while performing feature aggregation. This module consists of consecutive $3 \times 3$ separable convolutions with 128 outputs channels and a bilinear upsampling layer that upsamples by a factor of two. We add this mismatch correction module across the output of the second DPC and first LSFE branch, as well as between the two LSFE branches. We then upsample the outputs of each of these branches by a factor of four and subsequently concatenate them to yield 512 output filters which are then passed through a $3 \times 3$ separable convolution with $N_{\text{‘stuff’+‘thing’}}$ filters. Finally, this resulting tensor is upsampled by a factor of two and subsequently fed to a softmax layer which yields the semantic logits.

For training our semantic head, we minimize the weighted per-pixel log-loss~\cite{bulo2017loss} for a batch size $n$ given by
\begin{equation}\label{eq:Lsemantic}
\mathcal{L}_{semantic}(\Theta)=-\frac{1}{n}\sum \sum_{ij} w_{ij}log \; p_{ij}(p_{ij}^*),
\end{equation}
where $p_{ij}$ and $p_{ij}^*$ are the predicted and groundtruth class of pixel $(i,j)$ respectively. All convolutions in our semantic segmentation head are followed by iABN sync and Leaky ReLU activation function.

\subsection{Instance Segmentation Head}

Our instance segmentation head is based on the Mask R-CNN~\cite{he2017mask} framework, a widely used architecture that augments the object classification and bounding box regression heads in Faster R-CNN~\cite{ren2015faster} with a mask generation branch. Mask R-CNN consists of two stages: Regional Proposal Network~(RPN) and RoIAlign feature extraction. The RPN takes the 2-way FPN features as input and produces a set of possible bounding boxes known as candidates. Subsequently, RoIAlign extracts candidate-specific features used to simultaneously predict the class, bounding box, and instance segmentation mask of each instance candidate in two parallel branches. The first branch is a mask segmentation network that consists of four $3 \times 3$ separable convolutions with 256 output filters, followed by a $2 \times 2$ transposed convolution with stride 2 and 256 filters, and a $1 \times 1$ convolution having $N_{\text{‘thing’}}$ output channels. This branch generates a mask logit of dimension $28 \times 28$ for each considered class. The second branch performs bounding box regression and object classification simultaneously. It consists of two consecutive fully-connected layers with 1024 channels and an additional task-specific fully-connected layer with $4*N_{\text{‘thing’}}$ and $N_{\text{‘thing’}} + 1$ outputs for regression and classification respectively. As a result, the instance head is trained by minimizing the sum of each specific loss given by
\begin{equation}\label{eq:Linstance}
\mathcal{L}_{instance} = \mathcal{L}_{os} + \mathcal{L}_{op} + \mathcal{L}_{cls} + \mathcal{L}_{bbx} + \mathcal{L}_{mask},
\end{equation}
where $\mathcal{L}_{os}$, $\mathcal{L}_{op}$, $\mathcal{L}_{cls}$, $\mathcal{L}_{bbx}$, and  $\mathcal{L}_{mask}$ correspond to the object score, proposal, classification, bounding box, and mask segmentation losses respectively, as defined in~\cite{he2017mask}. Additionally, we replace all the standard convolutional layers, batch normalization, and ReLU activation with separable convolution, iABN sync, and Leaky ReLU respectively.

\subsection{Instance Tracking Head}

With the aim of tracking object instances across consecutive frames, we incorporate a novel tracking head in parallel to the semantic and instance heads. For this purpose, we leverage the multiple objects detected by the instance head, their RoIAlign features, predicted class, and masks logits, as input to our tracking head. As a first step, we employ mask pooling~\cite{porzi2019learning} to only consider the information related to each detected object from the RoIAlign features thereby eliminating background pixels. In this step, we downsample the mask logits by a factor of two using maxpooling to match the resolution of the RoIAlign output. Specifically, we exploit the instance segmentation mask as an attention mechanism, where we use the output obtained from our instance head during testing and the groundtruth instance mask during training. 

We obtain a 256-dimensional feature vector from pooling under the object instance mask. These features are subsequently passed through two consecutive fully-connected layers with $128$ and $32\times N_{\text{‘thing’}}$ outputs respectively. This yields an association vector $a_s^c$ for each segment candidate $s \in S$ for the duration of $t$ frames that are considered in the loss function. As a result, we enable the network to learn an embedding space where feature vectors of the same semantic class $c$ and track ID $\psi$ are mapped metrically close to each other, while the segments of different object instances are mapped distantly. This embedding space is generated by minimizing the batch hard triplet loss~\cite{hermans2017defense} across $t$ frames with margin $\alpha$ as
{\setlength{\mathindent}{0cm}
\setlength{\belowdisplayskip}{0pt}
\begin{align}\label{eq:Ltrack}
\begin{split}
\mathcal{L}_{track} = \frac{1}{|S|} \sum_{s \in S}  \mathop{\mathrm{max}} \Big( & \;  \underset{e \in S}{\mathop{\mathrm{max}}} \; || a_s^{c,\psi} - a_e^{c,\psi} || \\
& - \underset{e \in S}{\mathop{\mathrm{min}}} ||a_s^{c,\psi} - a_e^{c,\bar{\psi}} || + \alpha, 0 \Big)
\end{split}
\end{align}}

Finally, we reconstruct the track IDs of each object instance in the inference stage. The main idea here is to associate the instances of different time frames that belong to the same tracklet and assign a unique track ID to them. To this end, we only consider new instances that have classification confidence scores higher than a certain threshold $u_{s}$. Subsequently, we measure the association similarities by means of the Euclidean distances between the current predicted embedding vectors at frame $t$ and the embedding vectors of previous object instances. Similar to~\cite{voigtlaender2019mots}, we then use the Hungarian algorithm to associate the instances while only considering the most recent track IDs within in a specific window of time $N_T$. Furthermore, we create a new track ID when instances with high classification confidence scores are not associated with previous track IDs.

\subsection{MOPT Fusion Module}

In order to yield the panoptic tracking output, we adaptively fuse the logits from the three task-specific heads of our architecture in our MOPT fusion module. The MOPT output consists of pixel-level predictions that either belong to ‘stuff’ or ‘thing’ classes or take a void value. Moreover, pixels predicted as ‘thing’ classes will also include instance and tracking IDs. 

During inference, we feed a sequence of $t$ frames into our network which generates a set of \textit{track} IDs, a set of \textit{segment candidates} with predicted class, confidence score, bounding box and mask logits, and a $M$-channel \textit{semantic logits}, from the three task-specific heads. As a first step, we attach track IDs to the corresponding segment. Thereafter, we filter the segment candidates to select the object instances that have confidence scores greater than a given threshold $u_p$. Subsequently, we rank the segments that are selected by their confidence scores and upsample their masks to match the input image resolution. As there can be potential overlaps between the masks, we resolve such conflicts by only retaining the higher ranking logits, thereby generating a clean set $B_d$ of instance mask logits. We obtain the complementary set of semantic mask logits $B_s$ by first selecting the channels $m \in M$ that have high class prediction scores and retaining only the instance mask logits in the area inside the corresponding bounding box while ignoring the mask logits that are outside. Having both sets of mask logits for each segment, we fuse them into a single segment mask adaptively by computing the Hadamard product similar to~\cite{mohan20epsn} as
\begin{equation}
B = (\sigma(B_d) + \sigma(B_s)) \odot (B_d + B_s).
\end{equation}

Finally, we concatenate the segment mask logits $B$ with the ‘stuff’ logits and apply the argmax operation along the channel dimension. In order to generate the pixel-wise panoptic tracking output, we first fill an empty canvas with the predictions of the instance-specific ‘thing’ classes and subsequently, fill the empty regions with the ‘stuff’ class predictions that have an area greater than a set threshold $u_a$.

\section{Evaluation Metrics}

In order to facilitate quantitative performance evaluations of MOPT, we adapt the standard Panoptic Quality (PQ)~\cite{kirillov2018panoptic} metric that is used to measure the performance of panoptic segmentation to account for the incorrectly tracked objects in MOPT. First, we formally define the MOPT task. For a given set of $C$ semantic classes encoded by $\nset{C} \coloneqq \{0,..., C-1\}$, the goal of the MOPT task is to map each pixel $i$ of a given frame $I_k \in \{I_{0},...,I_{t-1}\}$ to a pair $(c_{i},\psi_{i})_k \in \nset{C} \times \nset{N} $, where $t$ is the total number of frames, $I_k$ is the $k^{th}$ frame of the sequence, $c_{i}$ represents the semantic class of pixel $i$, and $\psi_{i}$ represents its track ID. The track ID $\psi_{i}$ associates a group of pixels having the same semantic class but belonging to a different segment and it is unique for each segment throughout the sequence. Moreover, if $c_{i} \in \nset{C}^{st}$, then its corresponding track ID $\psi_{i}$ is irrelevant. Considering the aforementioned task description, we define our proposed soft Panoptic Tracking Quality (sPTQ) metric based on the following criteria that the metric should
\begin{enumerate*}[label=(\roman*)]
  \item reflect the segmentation quality of ‘stuff’ and ‘thing’ classes uniformly,
  \item account for consistent tracking of object segments across time,
  \item be interpretable and straightforward for easy implementation.
\end{enumerate*}

To compute the sPTQ metric, we first establish the correspondences between the predicted object segment $p \in \nset{P}$ and the groundtruth object segment $g \in \nset{G}$. Here, $\nset{P}$ and $\nset{G}$ are sets of all the predicted and groundtruth object segments respectively. In this step, we take advantage of the non-overlapping mask property inherited from the panoptic segmentation task. This property guarantees that at most one predicted segment can have an Intersection-over-Union (IoU) higher than 0.5 for a given groundtruth mask. This results in unique matching that significantly simplifies the correspondence step as opposed to using bipartite matching to deal with the overlaps between multiple predicted segments and the groundtruth mask. Next, for a given class $c \in \nset{C}$, we compute the set of true positives $TP_c$, false positives $FP_c$, and false negatives $FN_c$, corresponding to matched pair of segments, unmatched predicted segments, and unmatched groundtruth  segments respectively as
{\setlength{\mathindent}{0cm}
\setlength{\belowdisplayskip}{0pt}
\begin{align}
TP_c &= \{p_c \in \nset{P} \;|\; IoU(p_c, g_c) > 0.5, \; \myforall \;\; g_c \in \nset{G}\}, \\
FP_c &= \{p_c \in \nset{P} \;|\; IoU(p_c, g) <= 0.5, \; \forall \; g \in \nset{G}\}, \\
FN_c &= \{g_c \in \nset{G} \;|\; IoU(g_c, p) <= 0.5, \; \forall \; p \in \nset{P}\}.
\end{align}}

Subsequently, we assess the segment consistency across the frames by keeping track of all occurrences where the track ID prediction $\psi_{p}$ changes compared to the $\psi_{g}$ in previous frames that belong to the $TP_c$ set. We do so for each class separately by accumulating the IoU score for such a segment where the track ID correspondence is incorrect. We refer to this term as sIDS, a soft version of the IDS presented in \cite{voigtlaender2019mots}. IDS and sIDS are defined as
{\setlength{\mathindent}{0cm}
\begin{gather}
IDS_c = \{p \in TP_c \;|\; \psi_{p} \ne \psi_{g}\}, \\
sIDS_c = \{IoU(p, g) \;|\; (p, g) \in TP_c \land \psi_{p} \ne \psi_{g}\}, 
\end{gather}}
where for each object segment, sIDS ranges in the interval $[0,1]$ taking a maximum value $1$ if a mismatch of track IDs occur. Consequently, our proposed metric sPTQ for a given class $c$ is given by
\begin{equation}
sPTQ_c = \frac{\sum_{(p,g) \in TP_c }IoU(p,g) - \sum_{{s \in IDS_c}}{s}} {|TP_c|+\frac{1}{2}|FP_c|+\frac{1}{2}|FN_c|}.
\end{equation}
 
sPTQ is comprised of three parts. First, the averaged IoU of matched segments $\frac{1}{|TP_c|}\sum_{(p,g) \in TP_c }IoU(p,g)$ accounts for the correct predictions. Second, the averaged IoU of matched segments with track ID discrepancy $\frac{1}{|TP_c|}\sum_{s \in IDS_c}{s}$ penalizes the incorrect track predictions. Finally, $\frac{1}{2}|FP_c|+\frac{1}{2}|FN_c|$ which is added to the denominator, penalizes the segments without matches. Considering that for $c \in \nset{C}^{st}$, the average IoU of matched segments with track ID mismatches is always zero, we also propose a stricter version of $sPTQ_c$ called $PTQ_c$. In this evaluation measure, we replace $sIDS_c$ in $sPTQ_c$ with $IDS_c$ as
\begin{equation}
PTQ_c = \frac{\sum_{(p,g) \in TP_c }IoU(p,g) - |IDS_c| } {|TP_c|+\frac{1}{2}|FP_c|+\frac{1}{2}|FN_c|}.
\end{equation}

The overall performance can be measured using sPTQ and PTQ which is averaged over all the classes as
\begin{align}\setlength{\belowdisplayskip}{0pt}
sPTQ=\frac{1}{|\nset{C}|}\sum\limits_{c \in \nset{C}}sPTQ_c,\\
PTQ=\frac{1}{|\nset{C}|}\sum\limits_{c \in \nset{C}}PTQ_c.
\end{align}
sPTQ jointly measures the performance of ‘stuff’ and ‘thing’ segmentation, as well as the tracking of ‘thing’ instances, therefore we adopt this metric as the primary evaluation criteria for the MOPT task. 

\begin{table*}
\begin{center}
\footnotesize
\begin{tabular}{p{5.4cm}|P{1cm}P{1cm}|P{1cm}P{1cm}P{1cm}|P{1cm}P{1cm}P{0.8cm}}
\toprule
Network & sPTQ & PTQ & sMOTSA & MOTSA & MOTSP & Params. & FLOPs & Time \\
 & (\%) & (\%) & (\%) & (\%) & (\%) & (\si{\million}) & (\si{\billion}) & (\si{\milli\second}) \\
\noalign{\smallskip}\hline\hline\noalign{\smallskip}
Seamless~\cite{porzi2019seamless} + Track~R-CNN~\cite{voigtlaender2019mots} & $45.66$ & $45.28$ & $18.09$ & $23.79$ & $84.28$ & $79.91$ & $273.96$ & $115$ \\
EfficientPS~\cite{mohan20epsn} + Track~R-CNN~\cite{voigtlaender2019mots} & $46.68$ & $46.3$ & $18.09$ & $23.79$ & $84.28$ & $69.91$ & $232.80$ & $115$ \\
EfficientPS~\cite{mohan20epsn} + MaskTrack~R-CNN~\cite{luiten2019video} & $46.17$ & $45.99$ & $17.74$ & $22.82$ & $83.78$ & $120.57$ & $224.43$ & $117$ \\
\midrule
PanopticTrackNet (ours) & $\mathbf{47.27}$ & $\mathbf{46.67}$ & $\mathbf{20.32}$ & $\mathbf{26.48}$ & $\mathbf{85.74}$ & $\mathbf{45.08}$ & $\mathbf{167.40}$ & $\mathbf{114}$ \\
\bottomrule
\end{tabular}
\end{center}
\caption{Vision-based panoptic tracking results on Virtual KITTI 2 validation set. The baselines combine predictions from individual task-specific models and their inference time was computed considering that the task-specific models are run in parallel.}
\label{tab:baselineVKITTI}
\end{table*}

\begin{table*}
\begin{center}
\footnotesize
\begin{tabular}{p{6.6cm}|P{0.7cm}P{0.8cm}|P{1cm}P{1cm}P{1cm}|P{0.8cm}P{0.8cm}P{0.7cm}}
\toprule
Network & sPTQ & PTQ & sMOTSA & MOTSA & MOTSP & Params. & FLOPs & Time \\
 & (\%) & (\%) & (\%) & (\%) & (\%) & (\si{\million}) & (\si{\billion}) & (\si{\milli\second}) \\
\noalign{\smallskip}\hline\hline\noalign{\smallskip}
RangeNet++~\cite{milioto2019rangenet} + PointPillars~\cite{lang2019pointpillars} + Track~R-CNN~\cite{voigtlaender2019mots} & $42.22$ & $41.94$ & $15.72$ & $21.93$ & $73.39$ & $110.74$ & $695.51$ & $409$ \\
KPConv~\cite{thomas2019kpconv} + PointPillars~\cite{lang2019pointpillars} + Track~R-CNN~\cite{voigtlaender2019mots} & $46.04$ & $45.50$ & $17.94$ & $23.78$ & $75.28$ & $89.96$ & $438.34$ & $514$ \\
EfficientPS~\cite{mohan20epsn} + Track~R-CNN~\cite{voigtlaender2019mots} & $44.50$ & $43.96$ & $18.86$ & $24.12$ & $75.57$ & $79.02$ & $379.73$ & $148$ \\
EfficientPS~\cite{mohan20epsn} + MaskTrack~R-CNN~\cite{luiten2019video} & $44.03$ & $43.72$ & $18.1$ & $23.9$ & $74.96$ & $120.64$ & $445.90$ & $151$ \\
\midrule
PanopticTrackNet (ours) & $\mathbf{48.23}$ & $\mathbf{47.89}$ & $\mathbf{25.35}$ & $\mathbf{30.09}$ & $\mathbf{77.34}$ & $\mathbf{45.13}$ & $\mathbf{300.81}$ & $\mathbf{146}$ \\
\bottomrule
\end{tabular}
\end{center}
\caption{LiDAR-based panoptic tracking results on SemanticKITTI validation set. The baselines combine predictions from individual task-specific models and their inference time was computed considering that the task-specific models are run in parallel.}
\label{tab:baselineSKITTI}
\vspace{-0.3cm}
\end{table*}

%Additionally, we evaluate and report the standard panoptic segmentation, and multi-object tracking and segmentation metrics. To this end, we use true positives $TP$, false positives $FP$, false negatives $FN$, and intersection-over-union $IoU = TP/(TP + FP + FN)$ to compute Standard Panoptic Quality (PQ), Segmentation Quality (SQ), and Recognition Quality (RQ) as
%\begin{align}
%PQ &= \frac{\sum_{(p,g) \in TP} IoU(p,g)} {|TP|+\frac{1}{2}|FP|+\frac{1}{2}|FN|},\\
%SQ &= \frac{\sum_{(p,g) \in TP}IoU(p,g)}{|TP|},\\
%RQ &= \frac{|TP|}{|TP|+\frac{1}{2}|FP|+\frac{1}{2}|FN|}.
%\end{align}

%We report the same metrics specifically for ‘Stuff’ classes (PQ\textsuperscript{St}, SQ\textsuperscript{St}, RQ\textsuperscript{St}) and ‘thing’ classes (PQ\textsuperscript{Th}, SQ\textsuperscript{Th}, RQ\textsuperscript{Th}). Furthermore, we include the the following metrics to present a panoptic baseline: Average Precision (AP), mean Intersection-over-Union (mIoU) for both ‘stuff’ and ‘thing’ classes, and the inference time.

%Moreover, we report the following multi-object tracking and segmentation metrics proposed in \cite{voigtlaender2019mots}:
%\begin{gather}
%\widetilde{TP} = \sum_{h \in TP} IoU(h,c(h)),\\
%MOTSA = \frac{|TP|-|FP|-|IDS|}{|M|},\\
%sMOTSA = \frac{\widetilde{TP}-|FP|-|IDS|}{|M|},\\
%MOTSP = \frac{\widetilde{TP}}{|TP|},
%\end{gather}
%which evaluate the detection, instance segmentation and tracking quality of the segment predictions.

%%%%%%%%%%%%%%%%%%%%%%%%%%%%%%%%%%%%%%%%%%%%%%%%%%%%%%%%%%%%%%%%%%%%%%%%%%%%%%%%
\section{Experimental Evaluation}

In this section, we first introduce several strong baselines for the MOPT task that combine predictions from state-of-the-art panoptic segmentation and multi-object tracking models. We then present both quantitative and qualitative results of vision-based MOPT on Virtual~KITTI~2~\cite{cabon2020virtual} and LiDAR-based MOPT on SemanticKITTI~\cite{behley2020benchmark} datasets. We report results using the sPTQ and PTQ metrics as well as the standard multi-object tracking and segmentation metrics~\cite{voigtlaender2019mots} and panoptic segmentation metrics~\cite{kirillov2018panoptic} for completeness.

\begin{table*}
\begin{center}
\footnotesize
\begin{tabular}{p{3.5cm}|p{0.5cm}p{0.5cm}p{0.5cm}|p{0.5cm}p{0.5cm}p{0.5cm}|p{0.5cm}p{0.5cm}p{0.5cm}|p{0.5cm}p{0.6cm}}
\toprule
Network & PQ & SQ & RQ & PQ\textsuperscript{Th} & SQ\textsuperscript{Th} & RQ\textsuperscript{Th} & PQ\textsuperscript{St} & SQ\textsuperscript{St} & RQ\textsuperscript{St} & AP & mIoU \\
\noalign{\smallskip}\hline\hline\noalign{\smallskip}
Panoptic FPN~\cite{kirillov2019panoptic} & $46.7$ & $77.7$ & $57.9$ & $37.5$ & $81.6$ & $45.3$ & $50.2$ & $76.2$ & $62.7$ & $26.1$ & $53.0$ \\
UPSNet~\cite{xiong2019upsnet} & $47.4$ & $78.3$ & $58.5$ & $38.1$ & $82.3$ & $46.1$ & $50.9$ & $76.8$ & $63.2$ & $26.4$ & $52.8$ \\
Seamless~\cite{porzi2019seamless} & $48.6$ & $79.0$ & $59.7$ & $39.4$ & $83.6$ & $48.2$ & $52.1$ & $77.4$ & $64.1$ & $27.2$ & $56.6$ \\
\midrule
PanopticTrackNet (ours) & $\mathbf{50.3}$ & $\mathbf{80.4}$ & $\mathbf{60.5}$ & $\mathbf{41.7}$ & $\mathbf{84.9}$ & $\mathbf{49.3}$ & $\mathbf{53.5}$ & $\mathbf{78.7}$ & $\mathbf{64.7}$ & $\mathbf{28.0}$ & $\mathbf{57.3}$ \\
\bottomrule
\end{tabular}
\end{center}
\caption{Comparison of panoptic image segmentation performance on Virtual KITTI 2 validation set. Results in [\%]. Note that only PanopticTrackNet performs the MOPT task. The baselines only perform panoptic segmentation.}
\label{tab:benchVKITTI}
\end{table*}

\begin{table*}
\begin{center}
\footnotesize
\begin{tabular}{p{4.5cm}|p{0.5cm}p{0.5cm}p{0.5cm}|p{0.5cm}p{0.5cm}p{0.5cm}|p{0.5cm}p{0.5cm}p{0.5cm}|p{0.5cm}p{0.6cm}|p{0.8cm}}
\toprule
Network & PQ & SQ & RQ & PQ\textsuperscript{Th} & SQ\textsuperscript{Th} & RQ\textsuperscript{Th} & PQ\textsuperscript{St} & SQ\textsuperscript{St} & RQ\textsuperscript{St} & AP & mIoU & Time \\
\noalign{\smallskip}\hline\hline\noalign{\smallskip}
RangeNet++~\cite{milioto2019rangenet} + PointPillars~\cite{lang2019pointpillars} & $36.5$ & $73.0$ & $44.9$ & $19.6$ & $69.2$ & $24.9$ & $47.1$ & $75.8$ & $59.4$ & $12.1$ & $52.8$ & $409\si{\milli\second}$ \\
KPConv~\cite{thomas2019kpconv} + PointPillars~\cite{lang2019pointpillars} & $\mathbf{41.1}$ & $\mathbf{74.3}$ & $\mathbf{50.3}$ & $28.9$ & $69.8$ & $33.1$ & $\mathbf{50.1}$ & $\mathbf{77.6}$ & $\mathbf{62.8}$ & $16.1$ & $\mathbf{56.6}$ & $514\si{\milli\second}$\\
\midrule
PanopticTrackNet (ours) & $40.0$ & $73.0$ & $48.3$ & $\mathbf{29.9}$ & $\mathbf{76.8}$ & $\mathbf{33.6}$ & $47.4$ & $70.3$ & $59.1$ & $\mathbf{16.8}$ & $53.8$ & $\mathbf{146\si{\milli\second}}$ \\
\bottomrule
\end{tabular}
\end{center}
\caption{Comparison of panoptic LiDAR segmentation performance on SemanticKITTI validation set. Results in [\%]. Note that only PanopticTrackNet performs MOPT. The baselines are disjoint panoptic segmentation models (semantic model + instance model).}
\label{tab:benchKITTI}
\vspace{-0.3cm}
\end{table*}

\subsection{Training Methodology}

We train our PanopticTrackNet end-to-end by combining the three loss functions from each of the task-specific heads defined in Eq.~(\ref{eq:Lsemantic}), (\ref{eq:Linstance}), and (\ref{eq:Ltrack}). We minimize the final loss function given by $\mathcal{L} = \mathcal{L}_{semantic} + \mathcal{L}_{instance} + \mathcal{L}_{track}$. During training, we use mini-sequences of length $t_w$ as input. We experimentally compared the performance with varying $t_w$ lengths and identify that $t_w=3$ yields the best performance. We use $\alpha = 0.2$ in the batch hard loss from \eqref{eq:Ltrack}. We train our model using a multi-step learning rate schedule with an initial learning rate of $0.007$ and weight decay of $0.1$ at epochs $\{31K,38K\}$ and $\{100K,124K\}$ for SemanticKITTI and Virtual~KITTI~2 respectively. We train our model using SGD with a momentum of $0.9$ for 90 epochs. We reconstruct the track ID with a classification confidence $u_s=0.5$ and window of time $N_T=3$. In the MOPT fusion module, we set the confidence score threshold $u_p=0.5$ and minimum area threshold $u_a$ as $375$ and $32$ for Virtual~KITTI~2 and SemanticKITTI respectively.

%We construct PanopticTrackNet by adding a tracking head and integrating architecture parts of MOTSNet~\cite{porzi2019learning} and EfficientPS~\cite{mohan20epsn}. We compute the proposed experiments based on MOTSNet configuration for tracking prediction and EfficientPS configuration for panoptic segmentation. The ground truth masks are taken as inputs of the mask pool operation during training and the predicted mask logits are used during the test. Moreover, we downsample the mask logits using maxpool operation of factor two to fit the RoIAlign output.

\subsection{Baseline Models}

To the best of our knowledge, there are no methods thus far that jointly perform semantic segmentation, instance segmentation and instance tracking. Therefore, we provide several new baselines for MOPT by combining predictions from state-of-the-art task-specific models.

For vision-based MOPT, we provide several baselines: \{Seamless~\cite{porzi2019seamless} + Track~R-CNN~\cite{voigtlaender2019mots}\}, \{EfficientPS~\cite{mohan20epsn} + Track~R-CNN~\cite{voigtlaender2019mots}\}, and \{EfficientPS~\cite{mohan20epsn} + MaskTrack~R-CNN~\cite{luiten2019video}\}. EfficientPS is the current state-of-the-art as well as the most efficient panoptic segmentation network and Seamless is the previous state-of-the-art top-down network. EfficientPS uses a shared backbone with the 2-way FPN coupled with a novel semantic segmentation head and a Mask R-CNN based instance head, whose outputs are fused adaptively to yield the panoptic segmentation prediction. While the Seamless model employs a shared backbone with the standard FPN coupled with a DeepLab inspired semantic head and a Mask R-CNN based instance head, whose outputs are fused similar to the first panoptic network~\cite{kirillov2018panoptic}. Track R-CNN and Mask Track R-CNN both perform multi-object tracking and segmentation (MOTS). Track R-CNN is based on Mask R-CNN and incorporates an association head that relates object instances across different frames. Similar to the instance head of our PanopticTrackNet, the association head of Track R-CNN outputs an embedding vector for each detected object. Consequently, this network learns an embedding space where the resulting vectors of different instances are placed distantly and vectors of the same instance are placed closely. MaskTrack R-CNN is a more recent model that builds upon Mask R-CNN and leverages instances similarity across frames to infer an object track. To this end, the instance features from a single frame are stored in an external memory and updated across frames to compute a multi-class classification problem.

For LiDAR-based MOPT, we provide four baselines: \{RangeNet++~\cite{milioto2019rangenet} + PointPillars~\cite{lang2019pointpillars} + Track~R-CNN~\cite{voigtlaender2019mots}\}, \{KPConv~\cite{thomas2019kpconv} + PointPillars~\cite{lang2019pointpillars} + Track~R-CNN~\cite{voigtlaender2019mots}\}, \{EfficientPS~\cite{mohan20epsn} + Track~R-CNN~\cite{voigtlaender2019mots}\}, and \{EfficientPS~\cite{mohan20epsn} + MaskTrack~R-CNN~\cite{luiten2019video}\}. As there are no panoptic segmentation architectures that directly learn in the 3D domain, we combine predictions from individual state-of-the-art 3D semantic segmentation and 3D instance segmentation networks. KPConv directly operates on point clouds, whereas RangeNet++ operates on spherical projection of point clouds. PointPillars is a state-of-the-art 3D instance segmentation approach that employs PointNets to learn point cloud features organized as pillars which are then fed to a 2D CNN and SSD detector~\cite{liu2016ssd}. Similar to vision-based MOPT baselines, we employ Track R-CNN and MaskTrack R-CNN to obtain the tracking predictions.

\begin{figure*}
\centering
\footnotesize
\setlength{\tabcolsep}{0.05cm}
\setlength{\fboxsep}{0pt}
{\renewcommand{\arraystretch}{1.5}% for the vertical padding
\begin{tabular}{p{5.65cm} p{5.65cm} p{5.65cm}}
\arrayrulecolor[gray]{0.8}
\includegraphics[width=\linewidth]{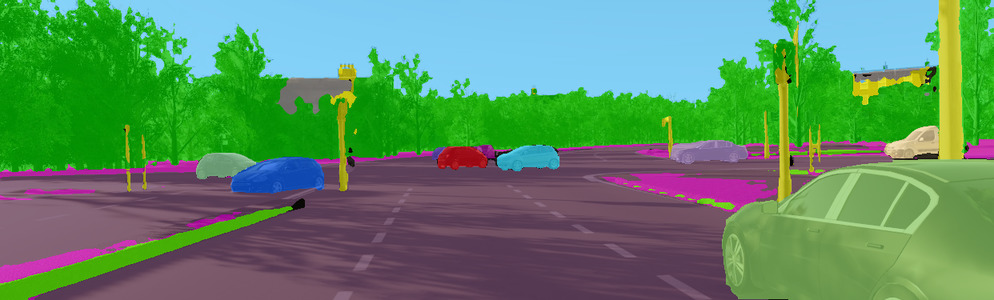} & \includegraphics[width=\linewidth]{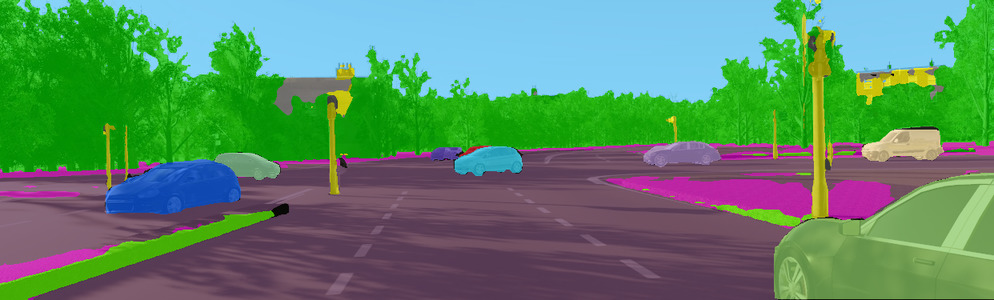} &
\includegraphics[width=\linewidth]{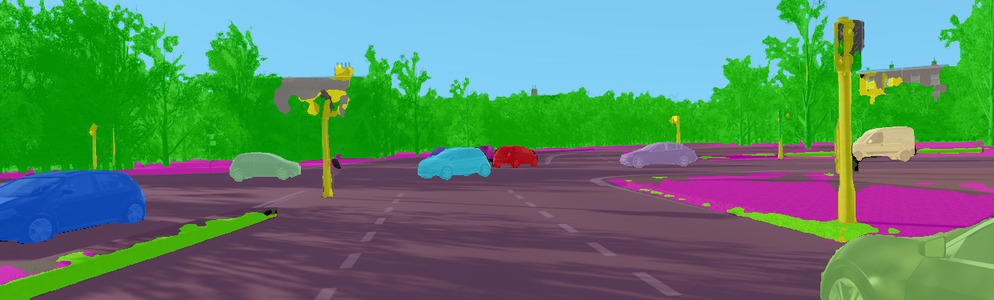} \\
\includegraphics[width=\linewidth]{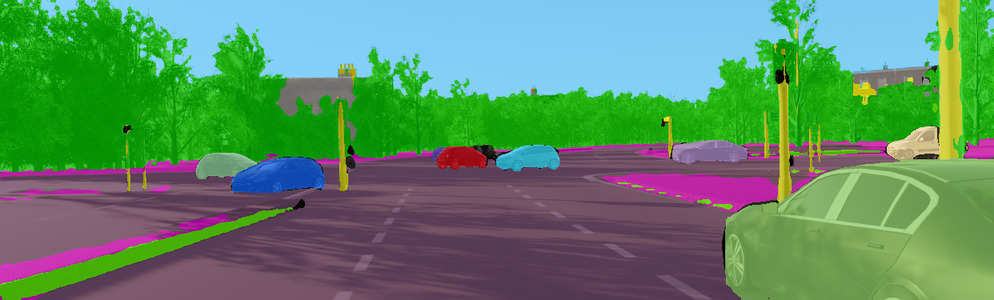} & 
\includegraphics[width=\linewidth]{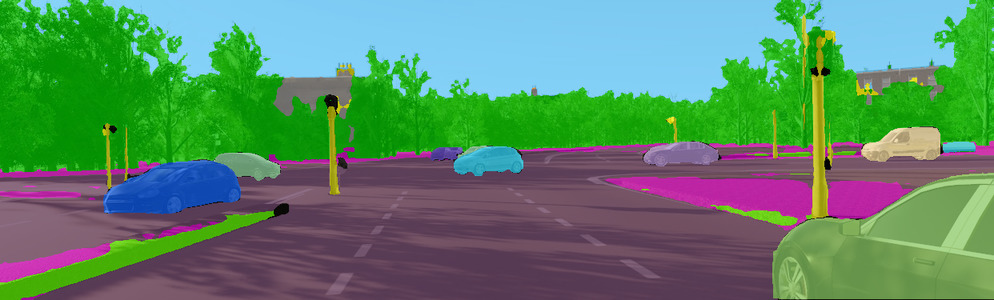} & 
\includegraphics[width=\linewidth]{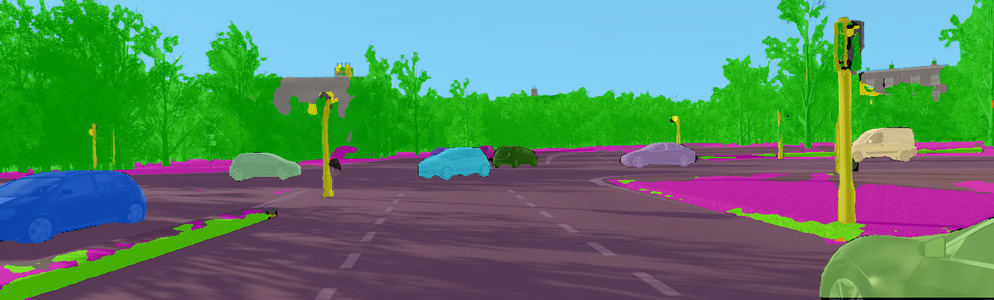} \\
\midrule[0.05cm]
\includegraphics[width=\linewidth]{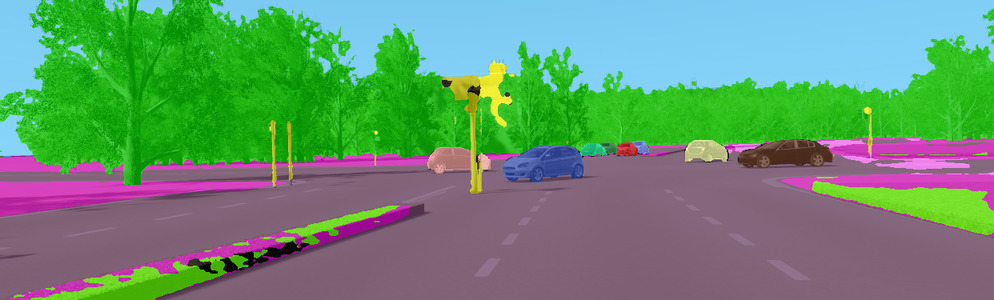} & \includegraphics[width=\linewidth]{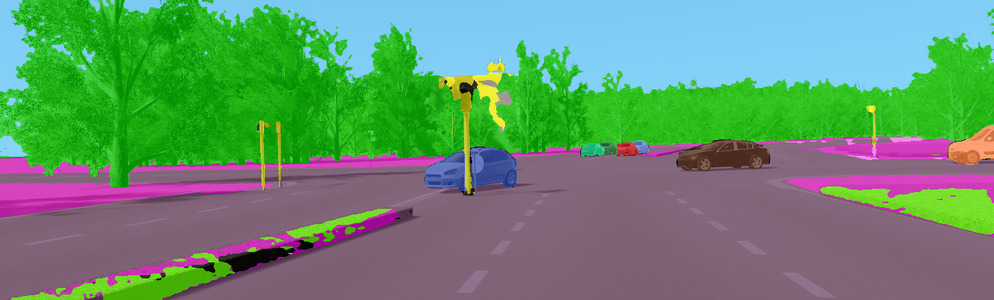} &
\includegraphics[width=\linewidth]{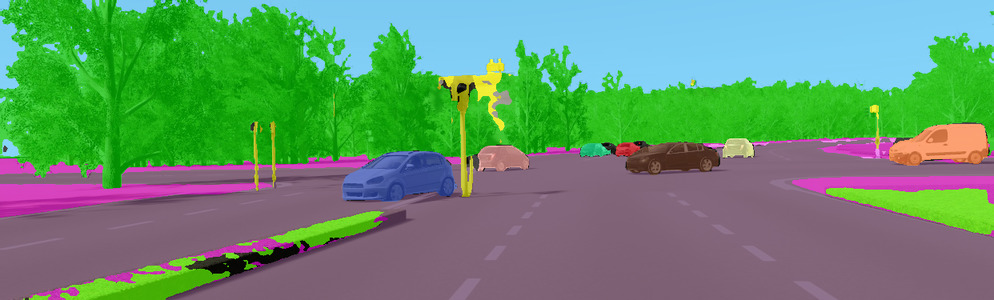} \\
\includegraphics[width=\linewidth]{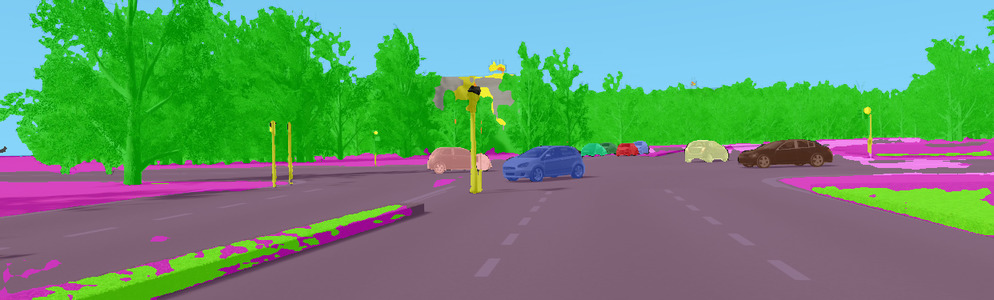} & 
\includegraphics[width=\linewidth]{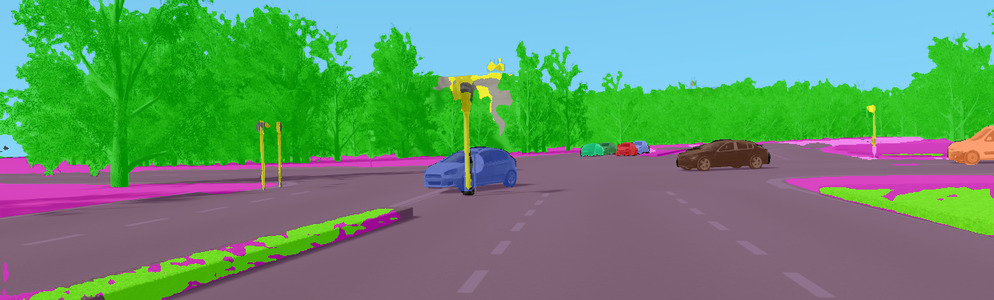} & 
\includegraphics[width=\linewidth]{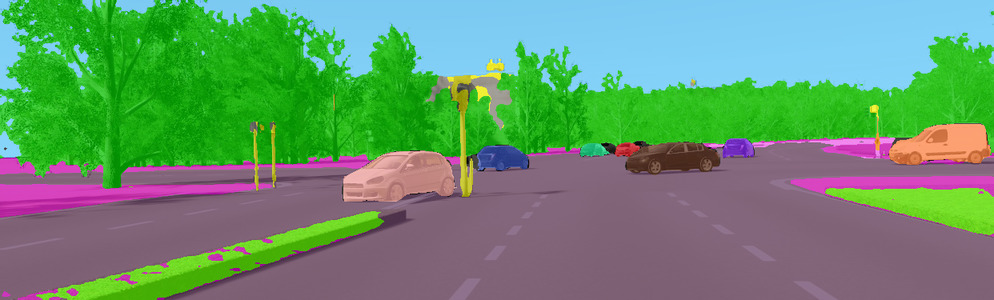} \\
\end{tabular}}
\caption{Qualitative comparisons of multi-object panoptic tracking (MOPT) from our proposed PanopticTrackNet (first and third rows) with \{EfficientPS + MaskTrack R-CNN\} (second and fourth rows) on Virtual KITTI 2 validation set. Each row shows the overlaid MOPT output of consecutive frames.}
\label{fig:qualitative}
\end{figure*}

\begin{figure*}
\centering
\footnotesize
\setlength{\tabcolsep}{0.05cm}
\setlength{\fboxsep}{0pt}
{\renewcommand{\arraystretch}{1.5}% for the vertical padding
\begin{tabular}{p{5.65cm} p{5.65cm} p{5.65cm}}
\arrayrulecolor[gray]{0.8}
\includegraphics[width=\linewidth, cfbox=lightgray 0.4pt 0pt 0pt]{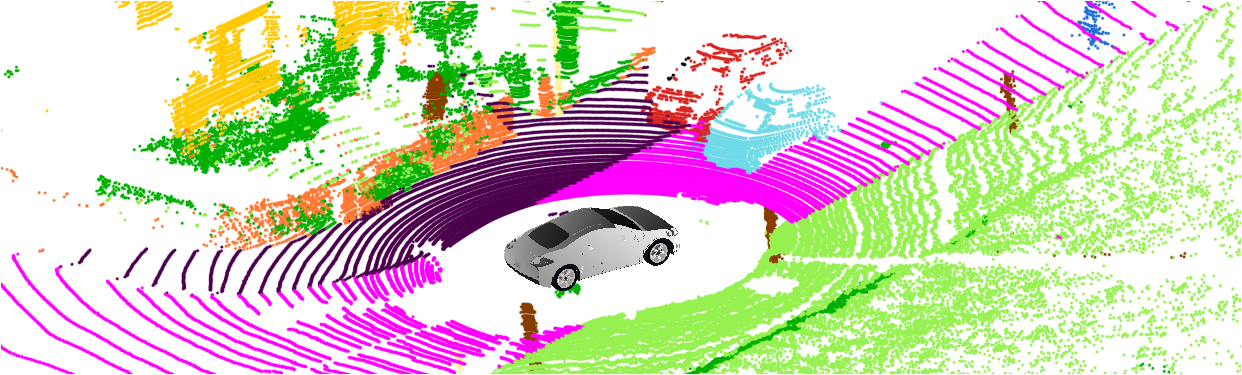} & \includegraphics[width=\linewidth, cfbox=lightgray 0.4pt 0pt 0pt]{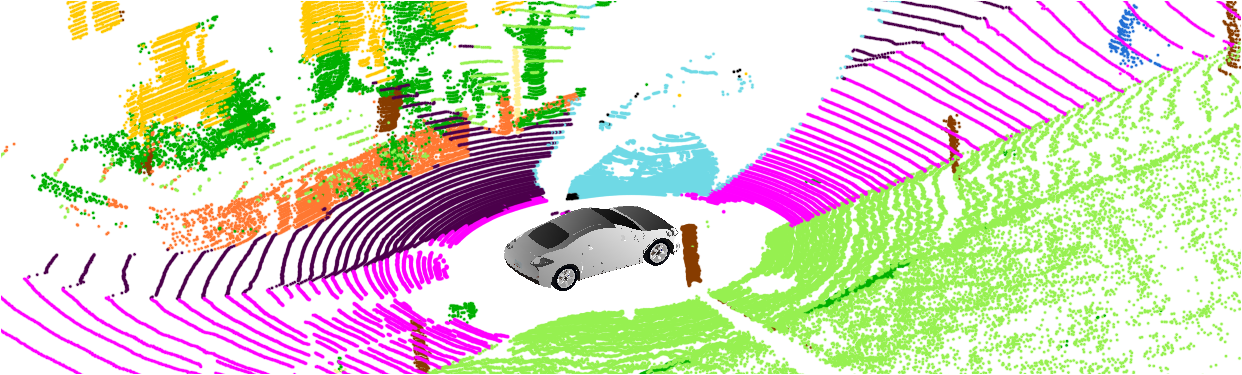} &
\includegraphics[width=\linewidth, cfbox=lightgray 0.4pt 0pt 0pt]{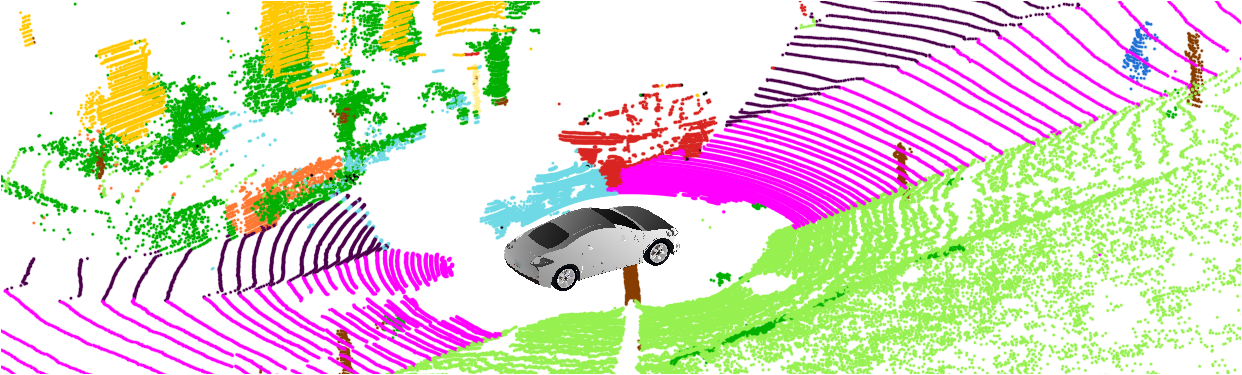} \\
\includegraphics[width=\linewidth, cfbox=lightgray 0.4pt 0pt 0pt]{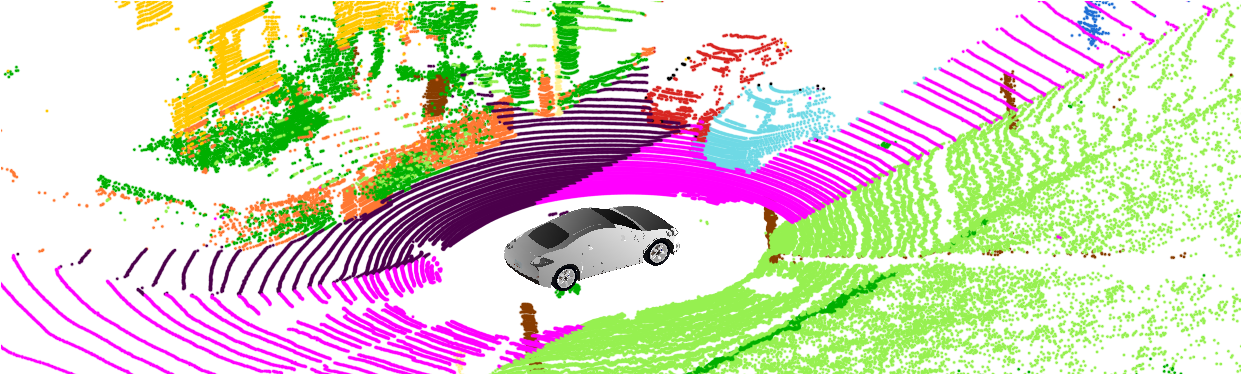} & 
\includegraphics[width=\linewidth, cfbox=lightgray 0.4pt 0pt 0pt]{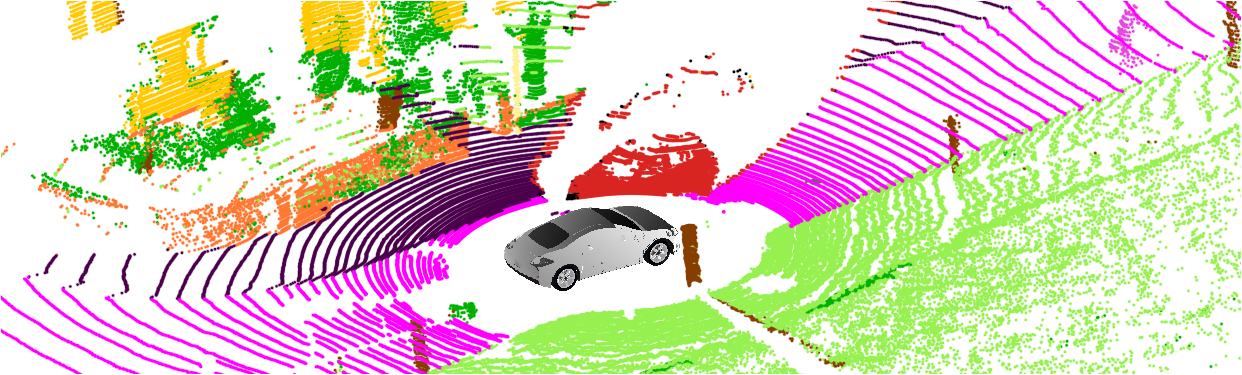} & 
\includegraphics[width=\linewidth, cfbox=lightgray 0.4pt 0pt 0pt]{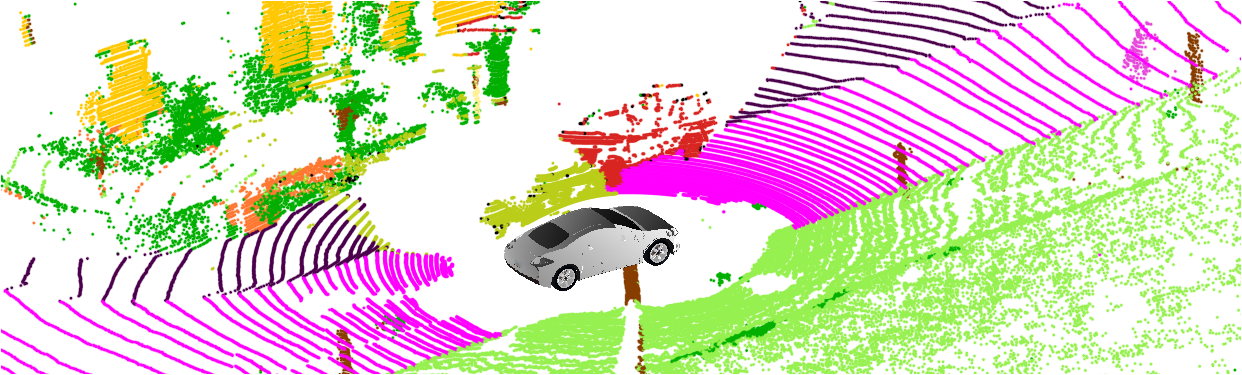} \\
\midrule[0.05cm]
\includegraphics[width=\linewidth, cfbox=lightgray 0.4pt 0pt 0pt]{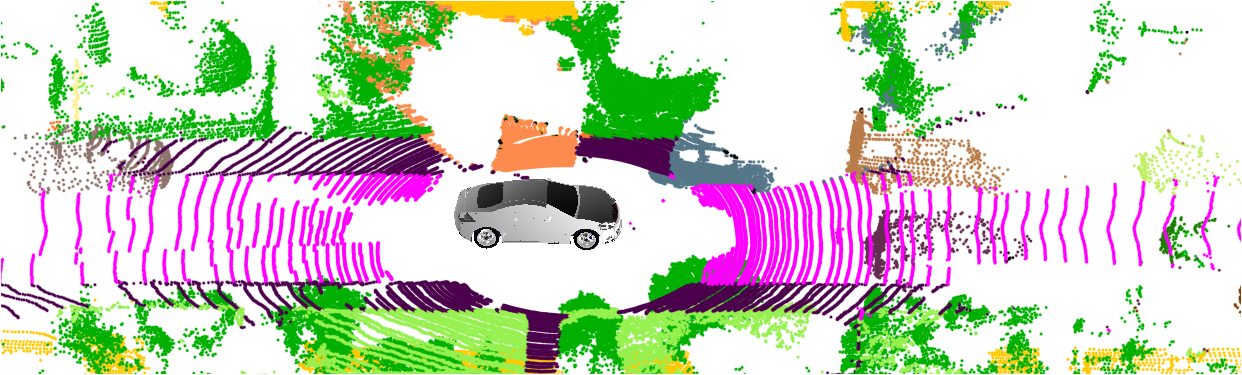} & \includegraphics[width=\linewidth, cfbox=lightgray 0.4pt 0pt 0pt]{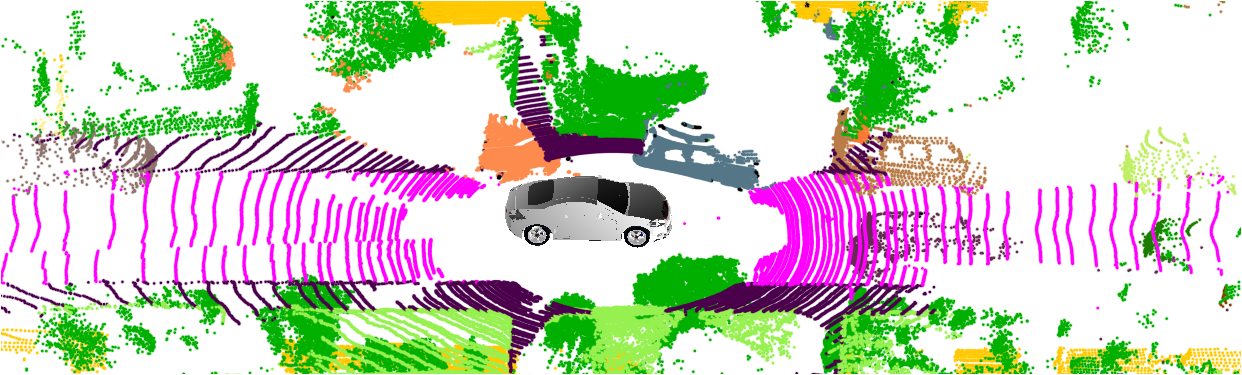} &
\includegraphics[width=\linewidth, cfbox=lightgray 0.4pt 0pt 0pt]{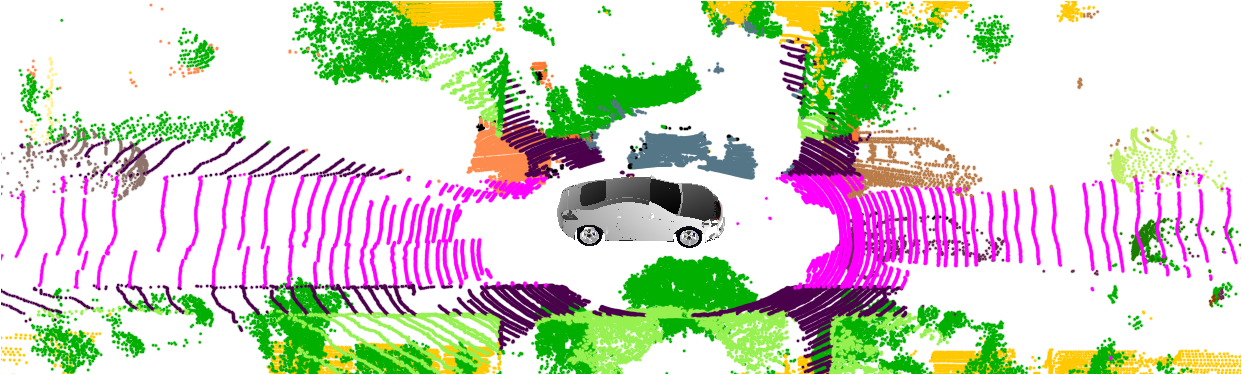} \\
\includegraphics[width=\linewidth, cfbox=lightgray 0.4pt 0pt 0pt]{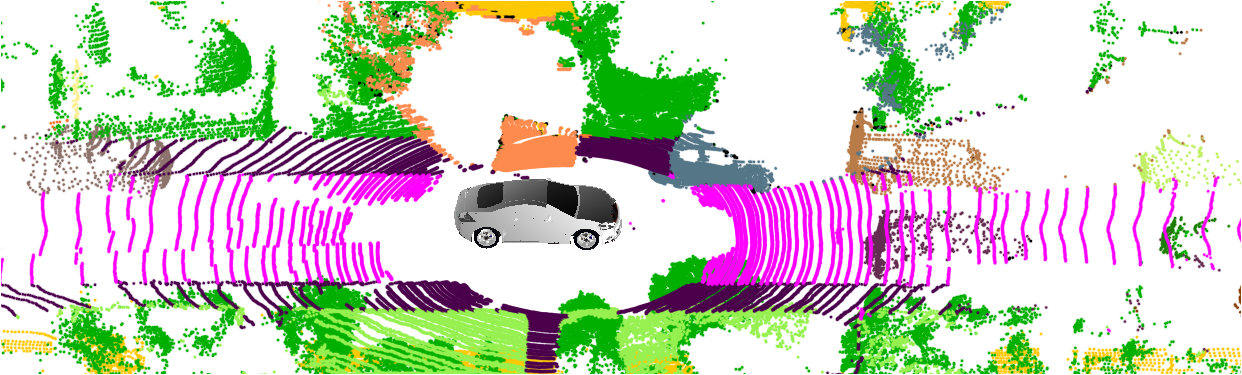} & 
\includegraphics[width=\linewidth, cfbox=lightgray 0.4pt 0pt 0pt]{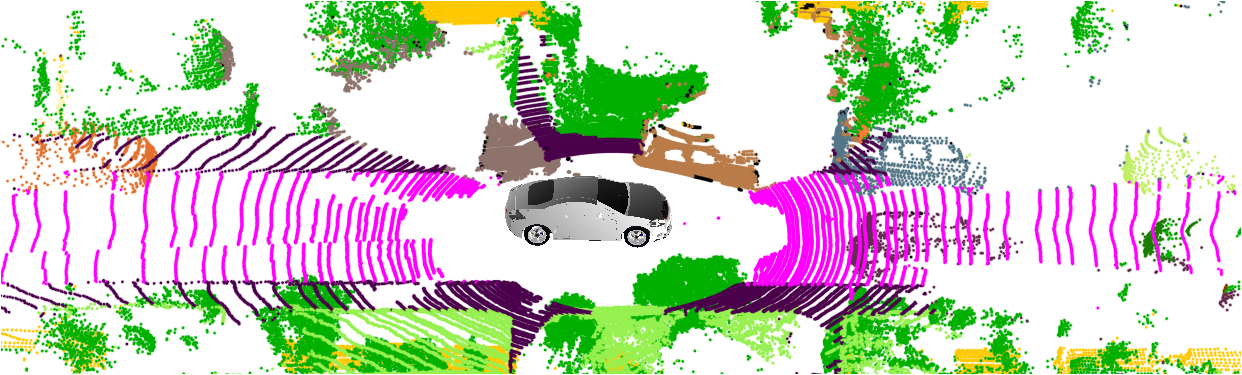} & 
\includegraphics[width=\linewidth, cfbox=lightgray 0.4pt 0pt 0pt]{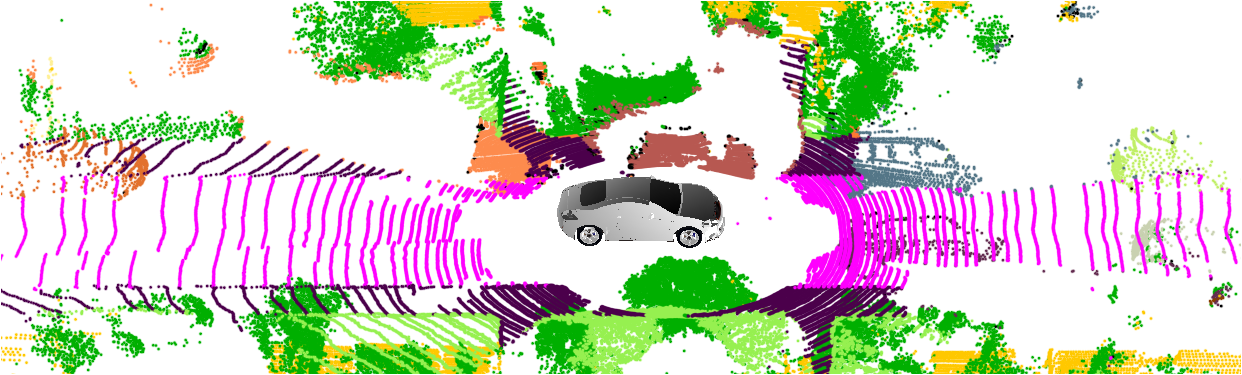} \\
\end{tabular}}
\caption{Qualitative comparisons of multi-object panoptic tracking (MOPT) from our proposed PanopticTrackNet (first and third rows) with \{EfficientPS + MaskTrack R-CNN\} (second and fourth rows) on the SemanticKITTI validation set. Each row shows the MOPT output of consecutive scans. A live demo can be seen at \url{http://rl.uni-freiburg.de/research/panoptictracking}.}
\label{fig:qualitative-lidar}
\vspace{-0.3cm}
\end{figure*}

\subsection{Quantitative Results}

We present results of vision-based MOPT on the Virtual KITTI 2 validation set in \tabref{tab:baselineVKITTI}. We observe that \{EfficientPS + Track~R-CNN\} achieves the highest performance among the baselines with a sPTQ score of $46.68\%$ and a sMOTSA score of $18.09\%$. While our proposed PanopticTrackNet model achieves a sPTQ score of $47.27\%$ and a sMOTSA score of $20.32\%$ which is an improvement of $0.58\%$ and $2.58\%$ respectively over the best performing baseline. Note that our proposed PanopticTrackNet architecture is built upon EfficientPS, therefore the improvement to the \{EfficientPS + Track~R-CNN\} model can be attributed to our track head, our adaptive MOPT fusion module that demonstrates a substantially better MOTS performance than Track R-CNN, and the end-to-end training of our unified model. The benefits of our unified model can also be observed in the number of parameters and FLOPs that it consumes, where it is almost twice as more efficient than the baselines.

In \tabref{tab:baselineSKITTI}, we present results of LiDAR-based MOPT on the SemanticKITTI validation set. The \{KPConv + PointPillars + Track R-CNN\} model which performs panoptic segmentation in the 3D domain achieves $46.02\%$ in sPTQ which is the highest among the baselines. However, it also has one of the largest inference times and FLOPs compared to the other baselines, which is a well known caveat of operating on point clouds directly. Nevertheless, our PanopticTrackNet achieves the highest sPTQ score of $48.23\%$ which is a large improvement of $2.19\%$ over the best performing baseline, while being almost four times faster in inference time. We also observe that our network achieves significantly higher scores in the MOTS metrics for both categories of baselines, the ones that directly operate on point clouds and the ones that operate on spherical projections. This improvement in performance and the computational efficiency truly demonstrates the capability of learning a single unified model for the MOPT task that is significantly more scalable.

Additionally, for completeness, we also present comparisons of panoptic segmentation performance on both Virtual KITTI 2 and SemanticKITTI validation sets in \tabref{tab:benchVKITTI} and \tabref{tab:benchKITTI} respectively. The results on Virtual KITTI 2 demonstrate that our proposed PanopticTrackNet achieves the state-of-the-art performance of $50.3\%$ in PQ. While the results on SemanticKITTI show that the \{KPConv + PointPillars\} model outperforms our PanopticTrackNet by $1.1\%$ in PQ due to the better performance on ‘stuff’ classes. However, our network achieves a higher performance on ‘things’ classes while being almost four times faster in inference time. These results demonstrate that our network that simultaneously performs panoptic segmentation and tacking outperforms specialized state-of-the-art panoptic segmentation models while being substantially more efficient as well as scalable.

\subsection{Qualitative Results}

In this section, we qualitatively evaluate the performance of our proposed PanopticTrackNet in comparison to the best performing baseline. We observe two complex scenarios of vision-based MOPT in \figref{fig:qualitative} where an object is occluded in one frame and reappears in the subsequent frames. Ideally, the same object should have the same track ID during the entire sequence. However, we show in the second and fourth rows that the baseline method loses track of the red and pink cars right after being occluded by other vehicles, whereas our network consistently tracks the objects. In \figref{fig:qualitative-lidar}, we present comparisons of LiDAR-based MOPT on SemanticKITTI in which we see that in the first example, both methods accurately identify the two cars denoted in blue and red. However, the red car gets occluded in the subsequent frame which causes the baseline to assign the same track ID to it as the adjacent car and when the car reappears in the next frame, the baseline assigns a new track ID to it which illustrates that the tracking is lost. Similarly, in the second example, the baseline loses track of multiple cars when they appear very close to each other. Nevertheless, our PanopticTrackNet yields consistent tracking and panoptic segmentation results in both these challenging scenes. This can be attributed to our mask-based tracking and inference head that also considers the predicted class and associates instances in the learned embedding space which enables it to consistently track objects even when their perspective changes.

%our network is able to track correctly the detected objects even when their forms change. These positive results can be explained by the track inference module used in our model that compares the current detections and previous track IDs in a time window  $N_t$  to avoid the generation of new track IDs after occlusion. Moreover, by using a mask-based tracking model that also includes the predicted class, our model is also able to track the objects when their perceptive form changes.

%%%%%%%%%%%%%%%%%%%%%%%%%%%%%%%%%%%%%%%%%%%%%%%%%%%%%%%%%%%%%%%%%%%%%%%%%%%%%%%%
\section{Conclusions}

In this work, we introduce and address a new perception task that we named Multi-Object Panoptic Tracking (MOPT) and the corresponding sPTQ metric for measuring the performance of our proposed task. MOPT unifies the conventionally disjoint problems of semantic segmentation, instance segmentation, and multi-object tracking into a single unified dynamic scene understanding task. This poses a set of unique challenges as well as gives ample opportunities to exploit complementary information from the sub-tasks. We proposed the novel PanopticTrackNet architecture that consists of a shared backbone with task-specific heads for learning to segment ‘stuff’ classes and ‘thing’ classes with temporally tracked instance masks. We demonstrated the performance of our model using two different modalities, namely vision-based MOPT on Virtual KITTI 2 and LiDAR-based MOPT SemanticKITTI. The results showed that our model exceeds the performance of several baselines comprised of state-of-the-art task-specific networks while being significantly faster and more efficient. We believe that these results demonstrate the viability of learning such scalable models for the MOPT task and opens avenues for future research.\looseness=-1

\section*{Acknowledgments}

This work was partly funded by the European Union's Horizon 2020 research and innovation program under grant agreement No 871449-OpenDR and by the Graduate School of Robotics in Freiburg. 

{\small
\bibliographystyle{ieee_fullname}
\bibliography{references.bib}

\begin{thebibliography}{10}\itemsep=-1pt

\bibitem{behley2020benchmark}
Jens Behley, Andres Milioto, and Cyrill Stachniss.
\newblock A benchmark for lidar-based panoptic segmentation based on kitti.
\newblock {\em arXiv preprint arXiv:2003.02371}, 2020.

\bibitem{bulo2017loss}
Samuel~Rota Bulo, Gerhard Neuhold, and Peter Kontschieder.
\newblock Loss max-pooling for semantic image segmentation.
\newblock In {\em Proc.~of the Conf.~on Computer Vision and Pattern
  Recognition}, pages 7082--7091, 2017.

\bibitem{cabon2020virtual}
Yohann Cabon, Naila Murray, and Martin Humenberger.
\newblock Virtual kitti 2.
\newblock {\em arXiv preprint arXiv:2001.10773}, 2020.

\bibitem{chen2018searching}
Liang-Chieh Chen, Maxwell Collins, Yukun Zhu, George Papandreou, Barret Zoph,
  Florian Schroff, Hartwig Adam, and Jon Shlens.
\newblock Searching for efficient multi-scale architectures for dense image
  prediction.
\newblock In {\em Advances in neural information processing systems}, pages
  8699--8710, 2018.

\bibitem{he2017mask}
Kaiming He, Georgia Gkioxari, Piotr Doll{\'a}r, and Ross Girshick.
\newblock Mask r-cnn.
\newblock In {\em Proc.~of the Int.~Conf.~on Computer Vision}, pages
  2961--2969, 2017.

\bibitem{hermans2017defense}
Alexander Hermans, Lucas Beyer, and Bastian Leibe.
\newblock In defense of the triplet loss for person re-identification.
\newblock {\em arXiv preprint arXiv:1703.07737}, 2017.

\bibitem{kirillov2019panoptic}
Alexander Kirillov, Ross Girshick, Kaiming He, and Piotr Doll{\'a}r.
\newblock Panoptic feature pyramid networks.
\newblock In {\em Proc.~of the Conf.~on Computer Vision and Pattern
  Recognition}, pages 6399--6408, 2019.

\bibitem{kirillov2018panoptic}
Alexander Kirillov, Kaiming He, Ross Girshick, Carsten Rother, and Piotr
  Doll{\'a}r.
\newblock Panoptic segmentation.
\newblock In {\em Proc.~of the Conf.~on Computer Vision and Pattern
  Recognition}, pages 9404--9413, 2019.

\bibitem{lang2019pointpillars}
Alex~H Lang, Sourabh Vora, Holger Caesar, Lubing Zhou, Jiong Yang, and Oscar
  Beijbom.
\newblock Pointpillars: Fast encoders for object detection from point clouds.
\newblock In {\em Proc.~of the Conf.~on Computer Vision and Pattern
  Recognition}, pages 12697--12705, 2019.

\bibitem{liu2016ssd}
Wei Liu, Dragomir Anguelov, Dumitru Erhan, Christian Szegedy, Scott Reed,
  Cheng-Yang Fu, and Alexander~C Berg.
\newblock Ssd: Single shot multibox detector.
\newblock In {\em European conference on computer vision}, pages 21--37, 2016.

\bibitem{luiten2019video}
Jonathon Luiten, Philip Torr, and Bastian Leibe.
\newblock Video instance segmentation 2019: A winning approach for combined
  detection, segmentation, classification and tracking.
\newblock In {\em Proc. of the IEEE International Conference on Computer Vision
  Workshops}, 2019.

\bibitem{luiten2020unovost}
Jonathon Luiten, Idil~Esen Zulfikar, and Bastian Leibe.
\newblock Unovost: Unsupervised offline video object segmentation and tracking.
\newblock {\em arXiv preprint arXiv:2001.05425}, 2020.

\bibitem{milioto2019rangenet}
Andres Milioto, Ignacio Vizzo, Jens Behley, and Cyrill Stachniss.
\newblock Rangenet++: Fast and accurate lidar semantic segmentation.
\newblock In {\em Proc.~of the IEEE/RSJ Int.~Conf.~on Intelligent Robots and
  Systems}, 2019.

\bibitem{mohan20epsn}
Rohit Mohan and Abhinav Valada.
\newblock Efficientps: Efficient panoptic segmentation.
\newblock {\em arXiv preprint arXiv:2004.02307}, 2020.

\bibitem{porzi2019seamless}
Lorenzo Porzi, Samuel~Rota Bulo, Aleksander Colovic, and Peter Kontschieder.
\newblock Seamless scene segmentation.
\newblock In {\em Proc.~of the Conf.~on Computer Vision and Pattern
  Recognition}, pages 8277--8286, 2019.

\bibitem{porzi2019learning}
Lorenzo Porzi, Markus Hofinger, Idoia Ruiz, Joan Serrat, Samuel~Rota Bul{\`o},
  and Peter Kontschieder.
\newblock Learning multi-object tracking and segmentation from automatic
  annotations.
\newblock {\em arXiv preprint arXiv:1912.02096}, 2019.

\bibitem{radwan2018multimodal}
Noha Radwan, Abhinav Valada, and Wolfram Burgard.
\newblock Multimodal interaction-aware motion prediction for autonomous street
  crossing.
\newblock {\em arXiv preprint arXiv:1808.06887}, 2018.

\bibitem{aiindex2019}
Erik Brynjolfsson Jack Clark John Etchemendy Barbara Grosz Terah Lyons James
  Manyika Saurabh~Mishra Raymond~Perrault, Yoav~Shoham and Juan~Carlos Niebles.
\newblock The ai index 2019 annual report.
\newblock Technical report, AI Index Steering Committee, Human-Centered AI
  Institute, Stanford University, Stanford, CA, December 2019.

\bibitem{ren2015faster}
Shaoqing Ren, Kaiming He, Ross Girshick, and Jian Sun.
\newblock Faster r-cnn: Towards real-time object detection with region proposal
  networks.
\newblock In {\em Advances in neural information processing systems}, pages
  91--99, 2015.

\bibitem{rota2018place}
Samuel Rota~Bul{\`o}, Lorenzo Porzi, and Peter Kontschieder.
\newblock In-place activated batchnorm for memory-optimized training of dnns.
\newblock In {\em Proc.~of the Conf.~on Computer Vision and Pattern
  Recognition}, pages 5639--5647, 2018.

\bibitem{sa2018weedmap}
Inkyu Sa, Marija Popovi{\'c}, Raghav Khanna, Zetao Chen, Philipp Lottes, Frank
  Liebisch, Juan Nieto, Cyrill Stachniss, Achim Walter, and Roland Siegwart.
\newblock Weedmap: a large-scale semantic weed mapping framework using aerial
  multispectral imaging and deep neural network for precision farming.
\newblock {\em Remote Sensing}, 10(9):1423, 2018.

\bibitem{schutt2019semantic}
Peer Sch{\"u}tt, Max Schwarz, and Sven Behnke.
\newblock Semantic interaction in augmented reality environments for microsoft
  hololens.
\newblock In {\em European Conference on Mobile Robots (ECMR)}, pages 1--6,
  2019.

\bibitem{tan2019efficientnet}
Mingxing Tan and Quoc~V Le.
\newblock Efficientnet: Rethinking model scaling for convolutional neural
  networks.
\newblock {\em arXiv preprint:1905.11946}, 2019.

\bibitem{thomas2019kpconv}
Hugues Thomas, Charles~R Qi, Jean-Emmanuel Deschaud, Beatriz Marcotegui,
  Fran{\c{c}}ois Goulette, and Leonidas~J Guibas.
\newblock Kpconv: Flexible and deformable convolution for point clouds.
\newblock In {\em Proc.~of the Int.~Conf.~on Computer Vision}, pages
  6411--6420, 2019.

\bibitem{valada2017deep}
Abhinav Valada and Wolfram Burgard.
\newblock Deep spatiotemporal models for robust proprioceptive terrain
  classification.
\newblock {\em The International Journal of Robotics Research},
  36(13-14):1521--1539, 2017.

\bibitem{valada2016convoluted}
Abhinav Valada, Ankit Dhall, and Wolfram Burgard.
\newblock Convoluted mixture of deep experts for robust semantic segmentation.
\newblock In {\em IEEE/RSJ International conference on intelligent robots and
  systems (IROS) workshop, state estimation and terrain perception for all
  terrain mobile robots}, 2016.

\bibitem{valada2018incorporating}
Abhinav Valada, Noha Radwan, and Wolfram Burgard.
\newblock Incorporating semantic and geometric priors in deep pose regression.
\newblock In {\em Workshop on Learning and Inference in Robotics: Integrating
  Structure, Priors and Models at Robotics: Science and Systems (RSS)}, 2018.

\bibitem{voigtlaender2019mots}
Paul Voigtlaender, Michael Krause, Aljosa Osep, Jonathon Luiten, Berin
  Balachandar~Gnana Sekar, Andreas Geiger, and Bastian Leibe.
\newblock Mots: Multi-object tracking and segmentation.
\newblock In {\em Proc.~of the Conf.~on Computer Vision and Pattern
  Recognition}, pages 7942--7951, 2019.

\bibitem{wang2019fast}
Qiang Wang, Li Zhang, Luca Bertinetto, Weiming Hu, and Philip~HS Torr.
\newblock Fast online object tracking and segmentation: A unifying approach.
\newblock In {\em Proc.~of the Conf.~on Computer Vision and Pattern
  Recognition}, pages 1328--1338, 2019.

\bibitem{wojek2010monocular}
Christian Wojek, Stefan Roth, Konrad Schindler, and Bernt Schiele.
\newblock Monocular 3d scene modeling and inference: Understanding multi-object
  traffic scenes.
\newblock In {\em European Conference on Computer Vision}, pages 467--481,
  2010.

\bibitem{xiong2019upsnet}
Yuwen Xiong, Renjie Liao, Hengshuang Zhao, Rui Hu, Min Bai, Ersin Yumer, and
  Raquel Urtasun.
\newblock Upsnet: A unified panoptic segmentation network.
\newblock In {\em Proc.~of the Conf.~on Computer Vision and Pattern
  Recognition}, pages 8818--8826, 2019.

\end{thebibliography}
}

\end{document}